# CROWD DENSITY ESTIMATION USING NOVEL FEATURE DESCRIPTOR

Adwan Alownie Alanazi[1] and Muhammad Bilal[2]

[1]Department of Computer Science & Software Engineering, University of Hail, Saudi Arabia
[2]International Gamming Lab, Italy

## ABSTRACT

*Crowd density estimation is an important task for crowd monitoring. Many efforts have been done to automate the process of estimating crowd density from images and videos. Despite series of efforts, it remains a challenging task. In this paper, we proposes a new texture feature-based approach for the estimation of crowd density based on Completed Local Binary Pattern (CLBP). We first divide the image into blocks and then re-divide the blocks into cells. For each cell, we compute CLBP and then concatenate them to describe the texture of the corresponding block. We then train a multi-class Support Vector Machine (SVM) classifier, which classifies each block of image into one of four categories, i.e. Very Low, Low, Medium, and High. We evaluate our technique on the PETS 2009 dataset, and from the experiments, we show to achieve 95% accuracy for the proposed descriptor. We also compare other state-of-the-art texture descriptors and from the experimental results, we show that our proposed method outperforms other state-of-the-art methods.*

## 1. INTRODUCTION

Crowd safety in pedestrian crowds is receiving great attention from the scientific community [46][47][48][49]. Mass festivals like sports, festivals, concerts, and carnivals, where a large amount of people gathers in a constrained environment [49][17], pose serious challenges to crowd safety [18][20]. In order to ensure safety and security of the participants, adequate safety measures should be adopted. Despite all safety measures, crowd disasters still occur frequently [21][24][25]. Therefore, crowd analysis is one of most important and challenging task in video surveillance.

The most important application of crowd analysis [22][23][26] can be used for crowd density estimation [27][28][29], crowd anomaly detection [12][13] and crowd flow segmentation [14][15] and tracking [16][30]. Among these application, crowd density estimation has received a significant importance from the research community [31][32]. In such public gathering, it is very crucial to estimate the density of crowd that can provide useful information. Acknowledging the importance of crowd density estimation, several attempts [33][39][45] have been made to tackle this problem with efficient algorithms [34][35]. In [1], the authors reports a most recent survey where the authors categorize and extensively evaluates different crowd density estimation approaches. From the extensive experimentations, the authors concluded that texture-based analysis [40][41][42] as compared to detection based methods [43][44][45] is the most robust and effective way of estimating crowd density [36][37][38]. Marana et al. [2, 3, 4, 5] was the first who showed the high density images exhibits fine texture and repetitive structures. They developed Gray Level Dependence Matrix (GLDM) [50] also known as Gray Level Co-occurrence Matrix (GLCM) that exploit texture features to estimate the crowd density. Fourier spectrum analysis and Minkowski fractal dimension (MFD) [6] were also used to estimate crowd density. These texture features were trained in [7] on different classifiers like Bayesian, support vector machine, Gaussian process regression, self-organizing neural network and fitting function-based approach. Rahmalan et al. [8] developed an approach based on new texture feature measure called Translation Invariant Orthonormal Chebyshev Moments (TIOCM). They compared TIOCM with GLDM and MFD and from the experimental results they showed that the performance of TIOCM is far better than MFD and almost same as GLCM, but GLCM takes more time for classifying images as compared to TIOCM. Local Binary Pattern (LBP), introduced by Ojala et al. [9], is new texture feature and extensively used for texture classification. LBP has the following advantages: (1) very easy to implement, (2) no need for pre-training, (3) invariance to illumination changes, and (4) low computational complexity. Due to these advantages, LBP is a preferred choice for many texture analysis applications.

Despite the advantages of LBP still have limitations, mostly the sensitivity to noise and scales. LBP is very sensitive to noise and is not capable to capture texture at different scales. Although some efforts have been made and various variants of LBP are proposed but all of these variants increase the computational complexity.

In this paper, we use the modified and extended version of LBP, called completed local binary pattern (CBLP) [10][51]   for crowd density estimation. For estimating crowd density estimation, we first divide the image into blocks. We then compute CBLP for each block and applied non-linear multi-class SVM for classification. We have performed experiments on different datasets and from the experimental results; we demonstrate the effectives CBLP features on crowd density estimation. The rest of the paper is organized as follows: Section 2 presents the related work; Section 3 elaborates our proposed method of features extraction; Section 4 presents experimental evaluation; and Section 5 concludes this paper.

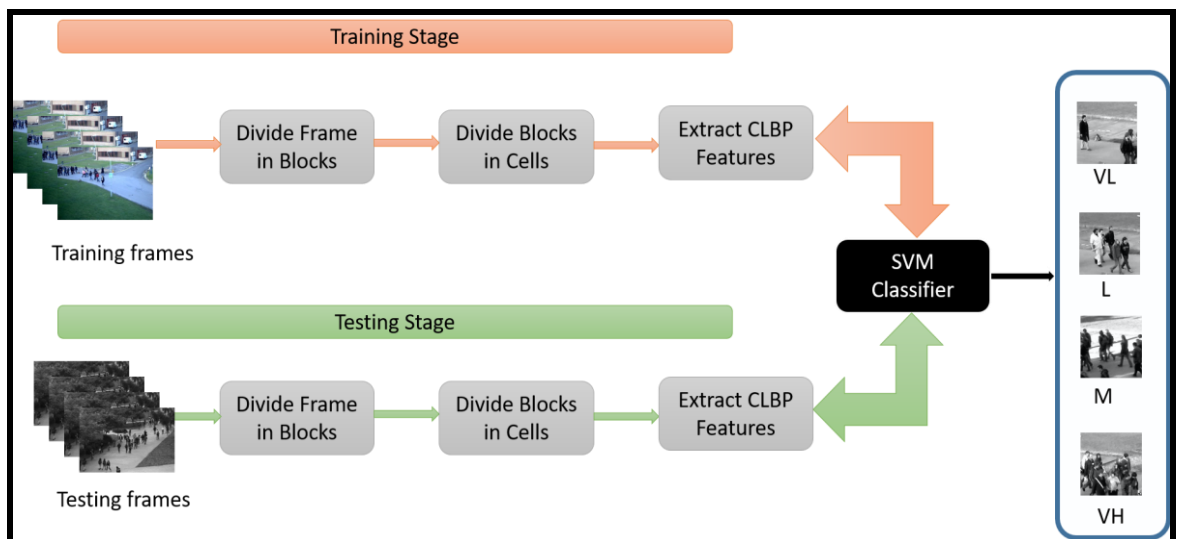

Figure 1. Flow diagram. The training frames are provided to train the neural network and testing frames are classified into four categories where VL represents very low, L represents low, M represents medium and VH represents very high density.

# 3. PROPOSED METHOD

We proposed a framework that will compute the CBLP features from the local regions of the image and classify each region of image into multiple density levels as shown in Figure 1. Our framework takes the input images and images are then divided into blocks of different crowd density levels. For extracting CBLP features, blocks are further again divided into overlapping cells. We compute CBLP for each cell and concatenated together to get a single feature vector for each block. Blocks are manually annotated to establish the ground truth according to the crowd density level as shown in Table I and in Figure 2. Figure 2 shows the sample of different crowd density levels. Our framework however has the following two advantages. (1) Overlapping cells within the block can encode more textures. (2) Division of frames into blocks can help in localization of crowded areas and reduce the effect of perspective.

Table 1: Classification of Different Crowd Density Levels

| Number of Persons | Density Level per meter square |
|---|---|
| Less than 7 | Very Low |
| Between 7 – 10 | Low |
| Between 11-16 | Medium |
| More than 26 | High |

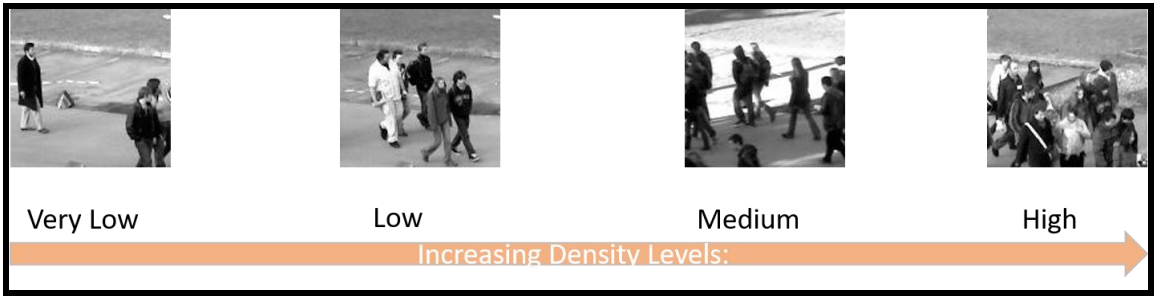

Figure 2: Increasing crowd density levels (image taken from PETS dataset)

## 3.1 Computation of completed local binary pattern (CLBP)

Completed Local Binary Patterns (CLBP) [10] composed of three variants of LBP- descriptors: CLBP C, LBP S and CLBP M. CLBP C include information on the center pixel, LBP S contain information about signed differences, and CLBP M contain the information about magnitudes of differences. The CLBP S descriptor is almost same as the original LBP descriptor; CLBP C set the thresholds value at the central pixel against the global average gray level value of the whole image and computed using the following formulation:

$$\text{CLBP\_C} = s\left(x_c - \frac{1}{MN}\sum_{i=1}^{N}\sum_{j=1}^{M}\mathbf{I}(i,j)\right)$$

Where M and N are the rows and column of the image and $\mathbf{I}(i,j)$ represents the current gray level value

CLBP M on the other hand performs a binary comparison between the absolute value of the difference between the central pixel and its neighbors and sets a global threshold to generate an traditional LBP code and formulated as follows

$$\text{CLBP\_M}_{r,p} = \sum_{n=0}^{p-1} s(|x_{r,p,n} - x_c| - \mu^g_{r,p}) 2^n$$

Where $\mu^g_{r,p}$ is a global threshold originally proposed by Guo *et al.* [11] is computed as:

$$\mu^g_{r,p} = \frac{\sum_{i=r+1}^{N-r} \sum_{j=r+1}^{M-r} \sum_{n=0}^{p-1} |x_{r,p,n}(i,j) - x(i,j)|}{(M-2r)(N-2r)p}$$

Due to the combination of three descriptors, CLBP has provided better texture classification performance than traditional LBP and other texture descriptors but the problem with CLBP is that it leads to high dimensionality due to the combination of these three descriptors.

## 4. Experimental evaluation

For the experimental evaluation we use Intel Core *i5*-2400@3.1GHz processor and 4GB RAM using MATLAB 2018a. We use *PETS*2009 dataset for the evaluation and comparison of other method with other methods. The resolution of frames in PETS2009 dataset is 768*X*576 with different viewpoints. In order to feed the frame to our model we first divide the analyze frame into set of blocks. Let frame $f_s$ is divided into *n* number off blocks. Let $B = \{b1, b2, ..., bn\}$. In this case, the size of each block is hyper-parameter. We extract texture features using different methods by using different block sizes and the results are reported in Figure 3. From the figure, it is obvious, that all texture descriptors are able to estimate crowd density with average accuracy of 70%. However, among all these methods, our proposed method out performs state of the art texture descriptors.

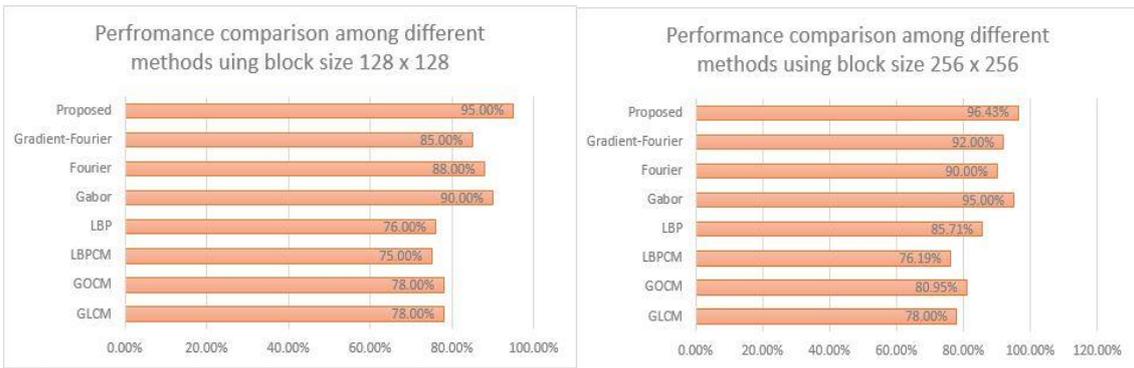

Figure 3: Comparison of different texture descriptors using different block sizes.

In order to rigorously analyse the performance of our proposed method, we build a confusion matrix and the results are reported in Table 2. In this experiment, we train SVM using one category images and test on the other category. From the table, it is obvious that SVM trained on Very Low category also classify images from Low category. The reason that there is very close boundary between these two categories. In the same way model trained on High category also classified some images from the medium.

Table 2: Confusion matrix for CLBP.

|  | **Very Low** | **Low** | **Medium** | **High** |
|---|---|---|---|---|
| **Very Low** | 89.4% | 10.5% | 0% | 0% |
| **Low** | 4.5% | 95.5% | 0% | 0% |
| **Medium** | 0% | 5.2% | 94.7% | 0% |
| **High** | 0% | 0% | 5.5% | 94.4% |

In the Figure 4, we report the qualitative results of some of samples frames. We first divide the images into grid of bocks and then into cells. We then compute CLBP feature descriptor and then feed each block to the SVM classifier. The model predicts the label for each block and the results are shown in the figure. From the figure, it is obvious that our model precisely predicts the label for each block.

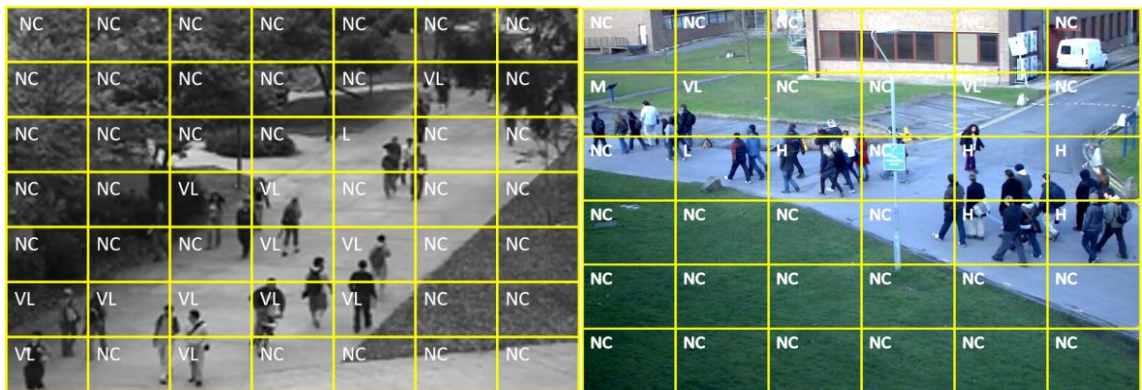

Figure 4: Sample frames with the predicted labels. The left image is sample frame taken from UCSD dataset while the right image is taken from PET2009 data set.

**5. Conclusions**

In this paper, we proposed an approach for crowd density estimation using our proposed features and non-linear SVM classifier. We demonstrated the capability of our approach in capturing the the dynamics of different classes by extracting these features. These features adopt the SVM to learn different classes. The main advantage of the proposed method is its simplicity and robustness.